\documentclass[review,12pt]{elsarticle} %Elsevier document class
\usepackage[a4paper,margin=2.2cm]{geometry} %Margin specifications
\usepackage[utf8]{inputenc} %Encoding for some characters
\usepackage[english]{babel}
\usepackage{caption}
\usepackage{subcaption}
\usepackage{array}
\usepackage[colorlinks=true]{hyperref} %Color hyperlinks
\usepackage{float}
\usepackage{multirow}
\usepackage{xurl}

\begin{document}
\begin{frontmatter}
	
	\title{Panoptic Segmentation Meets Remote Sensing}
	
	%% Authors and affiliations:
	\author[icc]{Osmar Luiz Ferreira de Carvalho}
	\ead{osmarcarvalho@ieee.org}
	\author[unb]{Osmar Abílio de Carvalho Júnior \corref{corresp}}
	\author[unb]{Cristiano Rosa e Silva}
	\ead{cristiano@dubbox.org}
	\author[unb]{Anesmar Olino de Albuquerque}
	\ead{anesmar@ieee.org}
	\author[unb]{Nickolas Castro Santana}
	\ead{nickolas.santana@unb.br}
	\author[icc]{Dibio Leandro Borges}
	\ead{dibio@unb.br}
	\author[unb]{Roberto Arnaldo Trancoso Gomes}
	\ead{robertogomes@unb.br}
	\author[unb]{Renato Fontes Guimarães}
	\ead{renatofg@unb.br}

	\affiliation[icc]{
		organization={University of Brasilia},
		department={Department of Computer Science,},
		addressline={Campus Universitario Darcy Ribeiro}, 
		city={Brasilia},
		postcode={70910-900}, 
		state={Federal District},
		country={Brazil}
	}
	\affiliation[unb]{
		organization={University of Brasilia},
		department={Departament of Geography,},
		addressline={Campus Universitario Darcy Ribeiro}, 
		city={Brasilia},
		postcode={70910-900}, 
		state={Federal District},
		country={Brazil}
	}

	\cortext[corresp]{Corresponding author: osmarjr@unb.br}
	
	\begin{abstract}
    Deep Learning (DL) methods achieved state-of-the-art results in remote sensing image segmentation studies with an increasing trend. Most studies focus on semantic and instance segmentation methods, with a research gap in panoptic segmentation. Panoptic segmentation combines instance and semantic predictions, allowing the detection of "things" (countable objects) and "stuff" (different backgrounds) simultaneously. Effectively approaching panoptic segmentation in remotely sensed data can be auspicious in many challenging problems since it allows continuous mapping and specific target counting. Several difficulties have prevented the growth of this task in remote sensing: (a) most algorithms are designed for traditional images, (b) image labeling must encompass "things" and "stuff" classes (being much more laborious), and (c) the annotation format is complex. Thus, aiming to solve and increase the operability of panoptic segmentation in remote sensing, this study has five objectives: (1) create a novel data preparation pipeline for the panoptic segmentation task using GIS tools, (2) propose a novel annotation conversion software to generate panoptic annotations in the COCO format automatically; (3) propose a novel dataset on urban areas, (4) modify and leverage the Detectron2 architecture and software for the task, and (5) evaluate semantic, instance, and panoptic metrics and present the difficulties of this task in the urban setting. We used an aerial image with a 0,24-meter spatial resolution in the city of Brasília, covering an area of 79,401 m$^2$. The annotations considered fourteen classes (three "stuff" and eleven "thing" categories). Our proposed pipeline considers three image inputs (original image, semantic image, and panoptic image). The proposed software uses these inputs alongside point shapefiles, creating samples at the centroid of each point shapefile with their corresponding annotations in the COCO format. The usage of points allows the researchers to choose samples in critical areas. Our study generated 3,400 samples with 512x512 pixel dimensions (3,000 for training, 200 for validation, and 200 for testing). The analysis used the Panoptic-FPN model with two backbones (ResNet-50 and ResNet-101), and the model evaluation considered three metric types (semantic metrics, instance metrics, and panoptic metrics). Regarding the main metrics, we obtained 93.865, 47.691, and 64.979 for the mean Intersection over Union, box Average Precision, and Panoptic Quality, respectively. Our study presents the first effective pipeline for panoptic segmentation and an extensive database for other researchers to use and deal with other data or related problems requiring a thorough scene understanding.
	\end{abstract}

	
\end{frontmatter}

\section{Introduction}
The increasing availability of satellite images alongside computational improvements makes the remote sensing field conducive to using deep learning (DL) techniques \citep{Ma2019Deep}. Unlike traditional machine learning (ML) methods for image classification that rely on a per-pixel analysis \citep{Maxwell2018Implementation, Shao2012Comparison}, DL enables the understanding of shapes, contours, textures, among other characteristics, resulting in better classification and predictive performance. In this regard, convolutional neural networks (CNNs) were a game-changing method in DL and pattern recognition because of its ability to process multi-dimensional arrays \citep{Lecun2015Deep}. CNNs apply convolutional kernels throughout the image resulting in feature maps, enabling low, medium, and high-level feature recognition (e.g., corners, parts of an object, and full objects, respectively) \citep{Nogueira2017Towards}. Besides, the development of new CNN architectures is a fast-growing field with novel and better architectures year after year, such as VGGnet \citep{Simonyan2015Very}, ResNet \citep{He2016Deep}, AlexNet \citep{Krizhevsky2017ImageNet}, ResNeXt \citep{Xie2017Aggregated}, Efficient-net \citep{Tan2019EfficientNet:}, among others.

\begin{figure}[H]
	\centering %
	\scriptsize %
	\includegraphics[width=0.7\columnwidth]{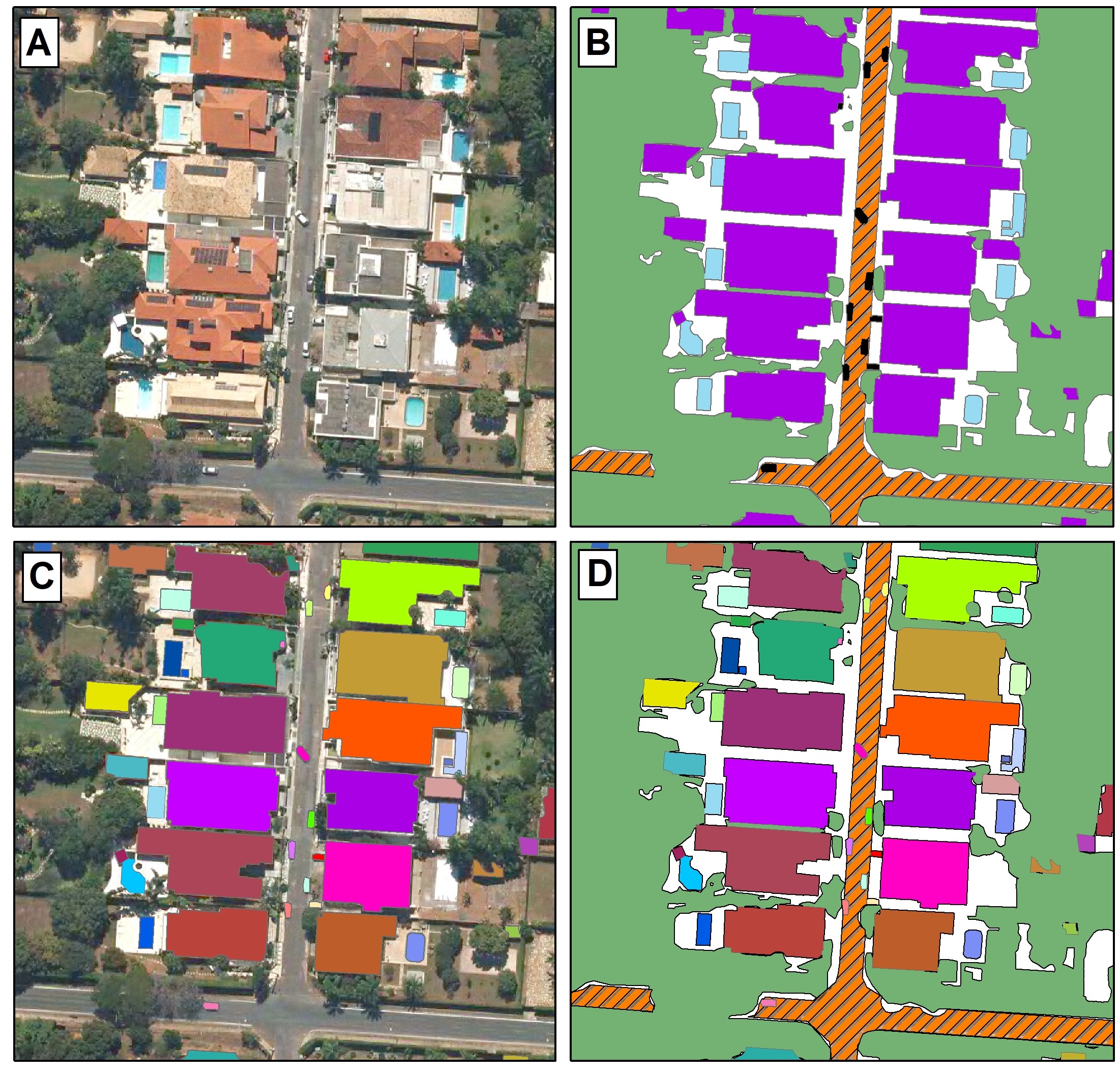}
	\caption{representation of the (A) Original image, (B) semantic segmentation, (C) instance segmentation, and (D) panoptic segmentation.}
	\label{fig:fig1}
\end{figure}

There are endless applications with CNN architectures, varying from single image classification to keypoint detection \citep{Dhillon2020Convolutional}. Nevertheless, there are three main approaches for image segmentation \citep{Hoeser2020Object, Ma2019Deep, Voulodimos2018Deep, Yuan2021review}: (1) semantic segmentation (Figure \ref{fig:fig1}B); (2) instance segmentation (Figure \ref{fig:fig1}C); and (3) panoptic segmentation (Figure \ref{fig:fig1}D). Semantic segmentation models perform a pixel-wise classification for a given input image \citep{Singh2020Semantic}, in which all elements belonging to the same class receive the same label. However, this method presents limitations for the recognition of individual elements, especially in crowded areas. On the other hand, instance segmentation generates bounding boxes (i.e., a set of four coordinates that delimits the object’s boundaries) and performs a binary segmentation mask for each element, enabling a distinct identification \citep{He2020Mask}. Nonetheless, instance segmentation approaches are restricted to objects, not covering background elements (e.g., lake, grass, roads). Most datasets adopt a terminology of “thing” and “stuff” categories to differentiate objects and backgrounds \citep{Cordts2016Cityscapes, Everingham2015Pascal, Geiger2013Vision, Lin2014Microsoft, Neuhold2017Mapillary}. The “thing” categories are often countable and present characteristic shapes, similar sizes, and identifiable parts (e.g., buildings, houses, swimming pools). Oppositely, “stuff” categories are usually not countable and amorphous (e.g., lake, grass, roads) \citep{Caesar2018COCO-Stuff:}. Thus, the panoptic segmentation task \citep{Kirillov2019Panoptic} aims to simultaneously combine instance and semantic predictions for classifying things and stuff categories, providing a more informative scene understanding.

Another crucial step for supervised DL is image annotation, which varies according to the segmentation task. Semantic segmentation is the most straightforward approach, requiring the original image and their corresponding ground truth images. The instance segmentation has a more complicated annotation style, which requires the bounding box information, the class identification, and the polygons that constitute each object. A standard approach is to store all of this information in the COCO annotation format \citep{Lin2014Microsoft}. Panoptic segmentation has the most complex and laborious format, requiring instance and semantic annotations. Therefore, the high complexity of panoptic annotations leads to a lack of remote sensing databases. Currently, panoptic segmentation algorithms are compatible with the standard COCO annotation format \citep{Kirillov2019Panoptic}. A significant advantage of using the COCO annotation format is compatibility with state-of-the-art software. Nowadays, Detectron2 \citep{Wu2019Detectron2} is one of the most advanced algorithms for instance and panoptic segmentation, and most research advances involve changes in the backbone structures, e.g., MobileNetV3 \citep{Howard2019Searching}, EfficientPS \citep{Mohan2021EfficientPS:}, Res2Net \citep{Gao2021Res2Net:}. Therefore, this format enables vast methodological advances. However, a big challenge in the application of remote sensing is the adaptation of algorithms to its peculiarities, which include the image format (e.g., GeoTIFF and TIFF) and the multiple channels (e.g., multispectral and time series), which differ from the traditional Red, Green, and Blue (RGB) images used in other fields of computer vision \citep{carvalho2021instance}.

\begin{figure}[H]
	\centering %
	\scriptsize %
	\includegraphics[width=0.5\columnwidth]{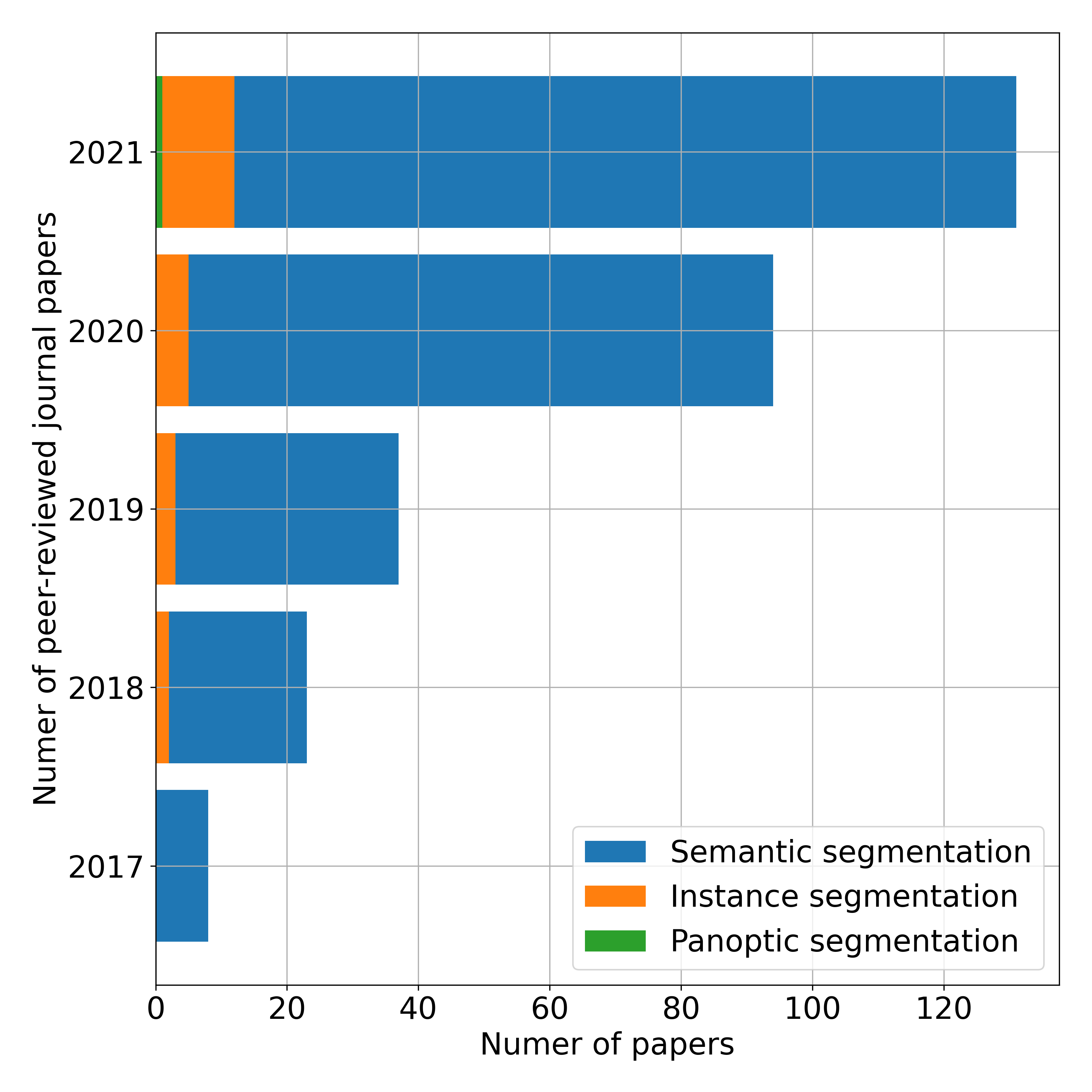}
	\caption{Temporal evolution of the number of articles in deep learning-based segmentation (semantic, instance and panoptic segmentation).}
	\label{fig:fig2}
\end{figure}

The increase in complexity among DL methods (panoptic segmentation $>$ instance segmentation $>$ semantic segmentation) reflects the frequency of peer-reviewed articles across each DL approach (Figure \ref{fig:fig2}). On the web of science database considering articles up to November 3, 2021, we evaluated four searches: (1) “remote sensing” AND “semantic segmentation” AND “deep learning”; (2) “remote sensing” AND “instance segmentation” AND “deep learning”; (3) “remote sensing” AND “panoptic segmentation” AND “deep learning” and (4) “panoptic segmentation”. Semantic segmentation is the most common approach using DL in remote sensing, with a total of 293 peer-reviewed articles. Instance segmentation has significantly fewer papers with a total of 22. On the other hand, panoptic segmentation has only one research published in remote sensing \citep{hua2021cascaded}, in which the authors used the DOTA \citep{xia2018dota}, UCAS-AOD \citep{liu2018linear}, and ISPRS-2D (\url{https://www2.isprs.org/commissions/comm2/wg4/benchmark/semantic-labeling/}) datasets, none of which are made for the panoptic segmentation task. Moreover, we found two other studies. The first focuses on change detection in building footprints using bi-temporal images \citep{Khoshboresh-Masouleh2021Building}, and the second use for different crops \citep{garnot2021panoptic}. Although both studies implement panoptic models, it does not use the main foundation of the panoptic proposition (since they do not consider multiple backgrounds), being very similar to an instance segmentation approach.

The absence of remote sensing panoptic segmentation research alongside databases for this task represents a substantial gap, mainly due to its high application potential. One of the notable drawbacks in the computer vision community regarding traditional images is the inference time, which exalts models like YOLACT and YOLACT++ \citep{Bolya2019YOLACT:, Bolya2020YOLACT++:} due to the ability to handle real data time even compromising the accuracy metrics a little. This problem is less significant in remote sensing as the image acquisition frequency is days, weeks, or even months, making it preferable to use methods that returns more information and higher accuracy rather than speed performance.

Moreover, the advancements of DL tasks are strictly related to the disposition of large publicly available datasets, being the case in most computer vision problems, mainly after the ImageNet dataset \citep{Deng2009ImageNet:}. These publicly available datasets encourage researchers to develop new methods to achieve ever-increasing accuracy and, consequently, new strategies that drive scientific progress. This phenomenon occurs in all tasks, shown by progressively better accuracy results in benchmarked datasets. What makes the COCO and other large datasets attractive to test new algorithms is: (1) an extensive number of images; (2) a high number of classes; and (3) the variety of annotations for different tasks. However, up until now, the publicly available datasets for remote sensing are insufficient. First, there are no panoptic segmentation datasets. Second, the instance segmentation databases are usually monothematic, as many building footprints datasets such as the SpaceNet competition \citep{Etten2018SpaceNet:}. 

A good starting point for a large remote sensing dataset would include widely used and researched targets, and the urban setting and its components is a very hot topic with many applications: road extraction \citep{Guo2020Multi-Scale, He2019Road,Kestur2018UFCN:, Lian2020DeepWindow:, Mokhtarzade2007Road, Senthilnath2020Deep, Wu2021Automatic, Xu2018Road}, building extraction \citep{Abdollahi2020Building, Bokhovkin2019Boundary, Griffiths2019Improving, Rastogi2020Automatic, Sun2021Semantic, Yi2019Semantic, Milosavljevic2020Automated}, lake water bodies \citep{Chen2018Encoder-Decoder, Guo2020Self-Supervised, Weng2020Water}, vehicle detection \citep{Ammour2017Deep, Audebert2017Segment-before-Detect:, Mou2018Vehicle}, slum detection \citep{Wurm2019Semantic}, plastic detection \citep{Jakovljevic2020Deep}, among others. Most studies address a single target at a time (e.g., road extraction, buildings), and panoptic segmentation would enable vast semantic information of images.

This study aims to solve these issues in panoptic segmentation for remote sensing images from data preparation up to implementation, presenting five contributions:

\begin{description}
  \item[$\bullet$] \textbf{BSB Aerial Dataset}: a novel dataset with a high amount of data and commonly used thing and stuff classes in the remote sensing community, suitable for semantic, instance, and panoptic segmentation tasks. 
  \item[$\bullet$] \textbf{Data preparation pipeline}: a novel method for preparing the ground truth data using commonly used GIS tools.
  \item[$\bullet$] \textbf{Annotation Converter}: a novel annotation converter from Geographic Information System (GIS) polygonal data into the panoptic, instance, and semantic segmentation format, which applies to other datasets.
  \item[$\bullet$] \textbf{Model modifications}: Panoptic-FPN leverage and modifications in the Detectron2 software for usability in remote sensing images applications (image format and number of channels).
  \item[$\bullet$] \textbf{Urban setting evaluation}: evaluation of semantic, instance, and panoptic segmentation metrics and evaluation of difficulties in the urban setting.
\end{description}

\section{Material and methods}
\begin{figure}[H]
	\centering %
	\scriptsize %
	\includegraphics[width=1\columnwidth]{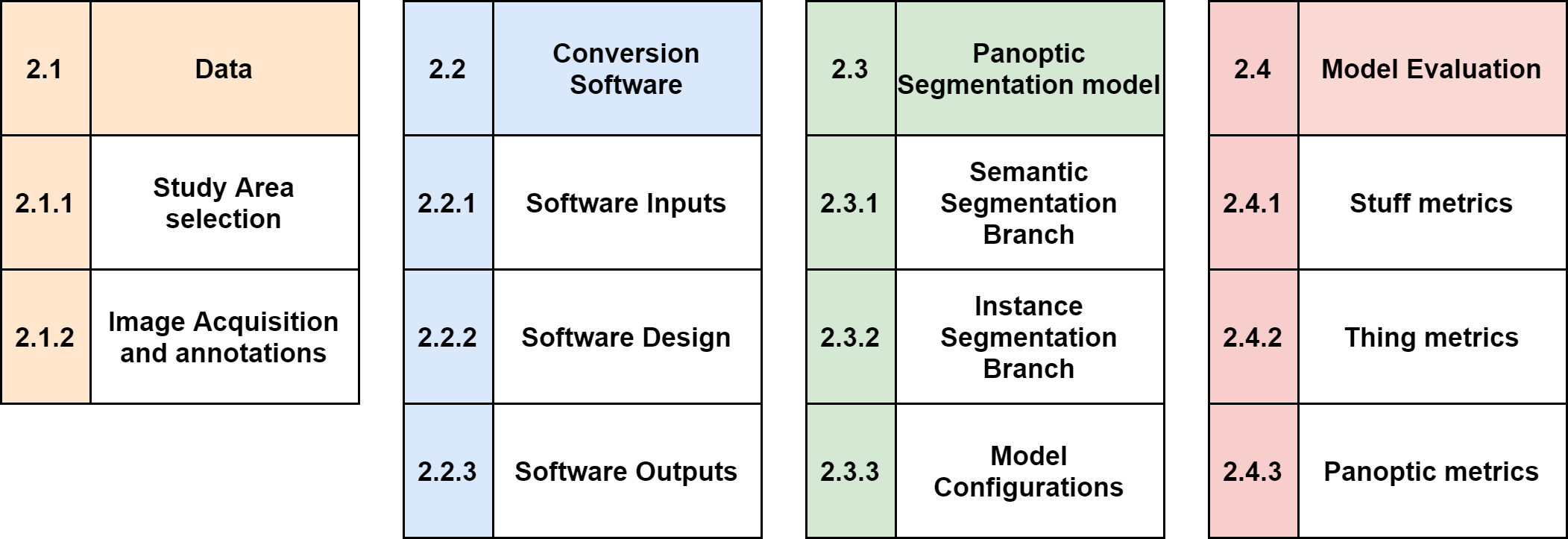}
	\caption{Methodological flowchart.}
	\label{fig:fig3}
\end{figure}

The present research had the following methodological steps (Figure \ref{fig:fig3}): (2.1) Data, (2.2) Conversion Software, (2.3) Panoptic Segmentation model, and (2.4) Model evaluation.

\subsection{Data}
\subsubsection{Study area selection}

\begin{figure}[!h]
	\centering %
	\scriptsize %
	\includegraphics[width=0.85\columnwidth]{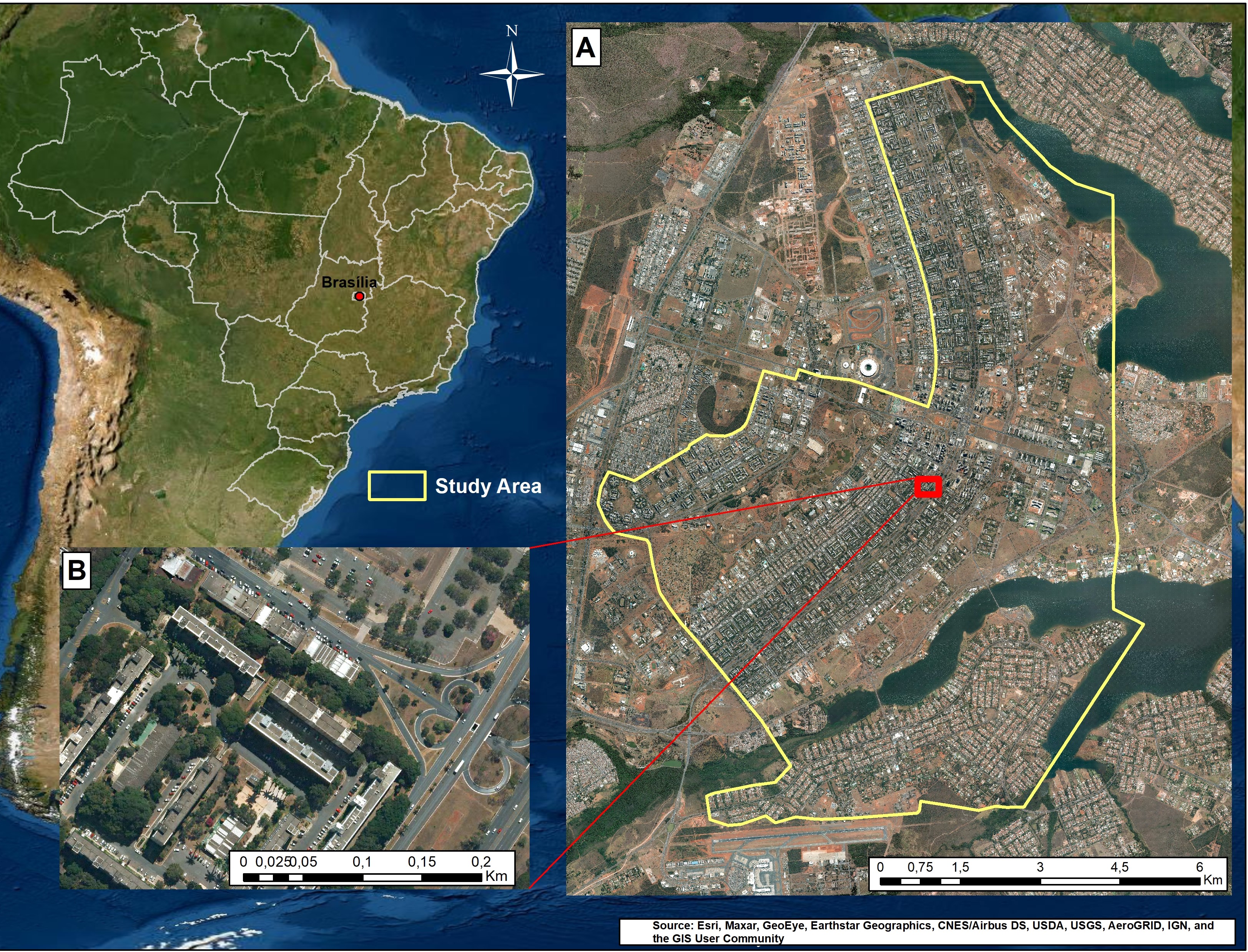}
	\caption{Study area.}
	\label{fig:fig4}
\end{figure}

The study area was the city of Brasília (Figure \ref{fig:fig4}), the capital of Brazil. Brasília was built and inaugurated in 1960 by President Juscelino Kubitschek to transfer the capital of Rio de Janeiro (in the coastal zone) to the country's central region, aiming at modernization and integrated development of the nation. The capital's original urban project was designed by the urban planner and architect Lúcio Costa, who modelled the city around Paranoá Lake with a top-view appearance of an airplane. The urban plan includes housing and commerce sectors around a series of parallel avenues 13 km long, containing zones dedicated to schools, medical services, shopping areas, and other community facilities. In 1988, United Nations Educational, Scientific and Cultural Organization (UNESCO) declared the city a World Heritage Site. 

The city presents suitable characteristics for DL tasks: (1) it is one of the few planned cities in the world presenting well-organized patterns, which eases the process of understanding each class; (2) the buildings are not high, which reduces occlusion and shadows errors due to the photographing angle; (3) the city contains organized portions of houses, buildings, and commerce, facilitating the annotation procedure; and (4) it has many socio-economical differences in many parts of the city, bringing information that might be useful to many other cities in the world.

\subsubsection{Image acquisition and annotations}

\begin{figure}[!htpb]
	\centering %
	\scriptsize %
	\includegraphics[width=1\columnwidth]{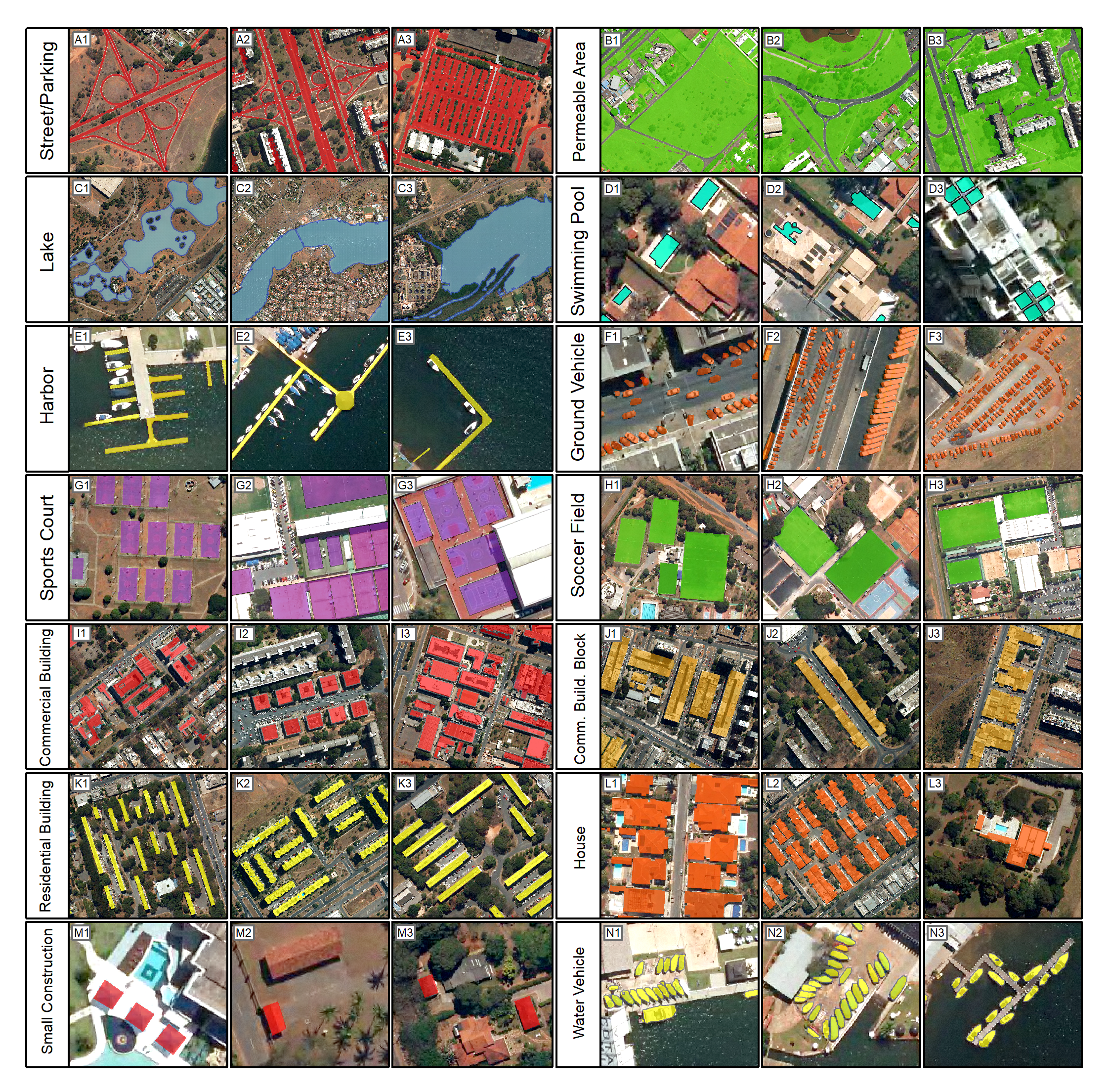}
	\caption{Three examples of each class from the proposed BSB Aerial Dataset.}
	\label{fig:fig5}
\end{figure}

The aerial images present the RGB channels and spatial resolution of 0.24 meters over Brasilia cover an area of 79,401 m$^2$, obtained by the Infraestrutura de Dados Espaciais do Distrito Federal (IDE/DF) (\url{https://www.geoportal.seduh.df.gov.br/geoportal/}). Annotations assign specific labels to targets of interest in the images. Specialists made vectorized annotations using the GIS software from traditional urban classes. Table \ref{tab:tab1} lists the panoptic categories considered in this study, and Table \ref{tab:tab2} lists the annotation pattern adopted for each class. The GIS specialists annotated fourteen different classes, in which three were “stuff” classes, and eleven were “things” classes (Figure \ref{fig:fig5}). The vehicles presented the most polygons (84,675), whereas the soccer fields had only 89. This imbalance among the different categories is widespread due to the nature of the urban landscape, e.g., there are more cars than soccer fields in cities. The understanding of this imbalance is an essential topic for investigating DL algorithms in the city setting. Since there is high variability in the permeable areas, we made a more generalized class considering all types of natural lands and vegetation, being the class with the highest number of annotated pixels (803,782,026).

\begin{table}[H]
\centering
\setlength\extrarowheight{-3pt}
\caption{category, numeric label, thing/stuff, and number of instances used in the BSB Aerial Dataset. The number of polygons in the stuff categories receive the '-' symbol since it is not relevant.}
\begin{tabular}{l|llll}
\hline
 Category            & Label & Thing/Stuff & Number of polygons  & Number of pixels \\
 \hline
 Background          &   0   &     -       &       -         &  112,497,999 \\
 Street              &   1   &  Stuff      &     -           &  167,065,309 \\
 Permeable Area      &   2   &  Stuff      &          -      &  803,782,026 \\
 Lake                &   3   &  Stuff      &           -     &  117,979,347 \\
 Swimming pool       &   4   &  Thing      &    4,835        &  3,816,585   \\
 Harbor              &   5   &  Thing      &    121          &  214,970     \\
 Vehicle             &   6   &  Thing      &    84,675       &  11,458,709  \\
 Boat                &   7   &  Thing      &    548          &  189,115     \\ 
 Sports Court        &   8   &  Thing      &    613          &  3,899,848   \\
 Soccer Field        &   9   &  Thing      &    89           &  3,776,903   \\
 Com. Buiding        &   10  &  Thing      &    3,796        &  69,617,961  \\
 Res. Buiding        &   11  &  Thing      &    1,654        &  8,369,418   \\
 Com. Building Block &   12  &  Thing      &    201          &  30,761,062  \\
 House               &   13  &  Thing      &    5,061        &  42,528,071  \\
 Small Construction  &   14  &  Thing      &    4,552        &  2,543,032   \\
 \hline
\end{tabular}
\label{tab:tab1}
\end{table}

\begin{table}[H]
\centering
\setlength\extrarowheight{-3pt}
\caption{Annotation pattern for each class in the BSB Aerial Dataset.}
\begin{tabular}{l|l}
\hline
 Category            & Annotation pattern                                                \\
 \hline 
 Background          &   Unlabeled pixels                                            \\
 Street              &   Visible asphalt areas                                           \\
 Permeable Area      &   Natural soil and vegetation (e.g., trees, land, grass)               \\
 Lake                &   Natural water bodies                             \\
 Swimming pool       &   Swimming pool polygons                                     \\
 Harbor              &   Harbor polygons                                            \\
 Vehicle             &   Ground vehicle polygons (e.g., car, bus, truck)        \\
 Boat                &   Boat polygons (in water and in land)                         \\ 
 Sports Court        &   Sports court polygons                         \\
 Soccer Field        &   Soccer field polygons                             \\
 Com. Buiding        &   Commercial building rooftop polygons                                \\
 Res. Buiding        &   Residential building rooftop polygons                               \\
 Com. Building Block &   Commercial building block rooftops polygons                          \\
 House               &   House-like polygons with area $>$ 80m$^2$         \\
 Small Construction  &   House-like polygons with area $<$ 80m$^2$        \\
 \hline
\end{tabular} 
\label{tab:tab2}
\end{table}

\subsection{Conversion Software}
DL methods require extensive collections of annotated images with different object classes for training and evaluation. Different open-sourced annotation software has been proposed, containing high-efficiency tools for the creation of polygons and bounding boxes, such as Labelme \citep{Russell2008LabelMe:, Torralba2010LabelMe:}, LabelImg (\url{https://github.com/tzutalin/labelImg}), Computer Vision Annotation Tool (CVAT) \citep{Sekachev2019Computer}, RectLabel (\url{https://rectlabel.com}), Labelbox (\url{https://labelbox.com}), and Visual Object Tagging Tool (VoTT) (\url{https://github.com/microsoft/VoTT}).However, the elaboration of annotations in remote sensing differs from other computer vision procedures that use traditional photographic images remote sensing annotations, containing some particularities, such as georeferencing, projection, multiple channels, and GeoTIFF files. Thus, there is a gap in specific annotation tools for remote sensing. In this context, a powerful solution for expanding the terrestrial truth database for deep learning is to take advantage of the extensive mapping information stored in a GIS database. Besides, GIS programs already have several editing, and manipulation tools developed and improved for geolocated data. Recently, a specific annotation tool for remote sensing is the LabelRS based on ArcGIS \citep{Li2021LabelRS:}, considering semantic segmentation, object detection, and image classification. However, LabelRS is based on ArcPy scripts dependent on ArcGIS, not fully open-source, and does not operate with panoptic annotations.

\begin{figure}[!h]
	\centering %
	\scriptsize %
	\includegraphics[width=0.85\columnwidth]{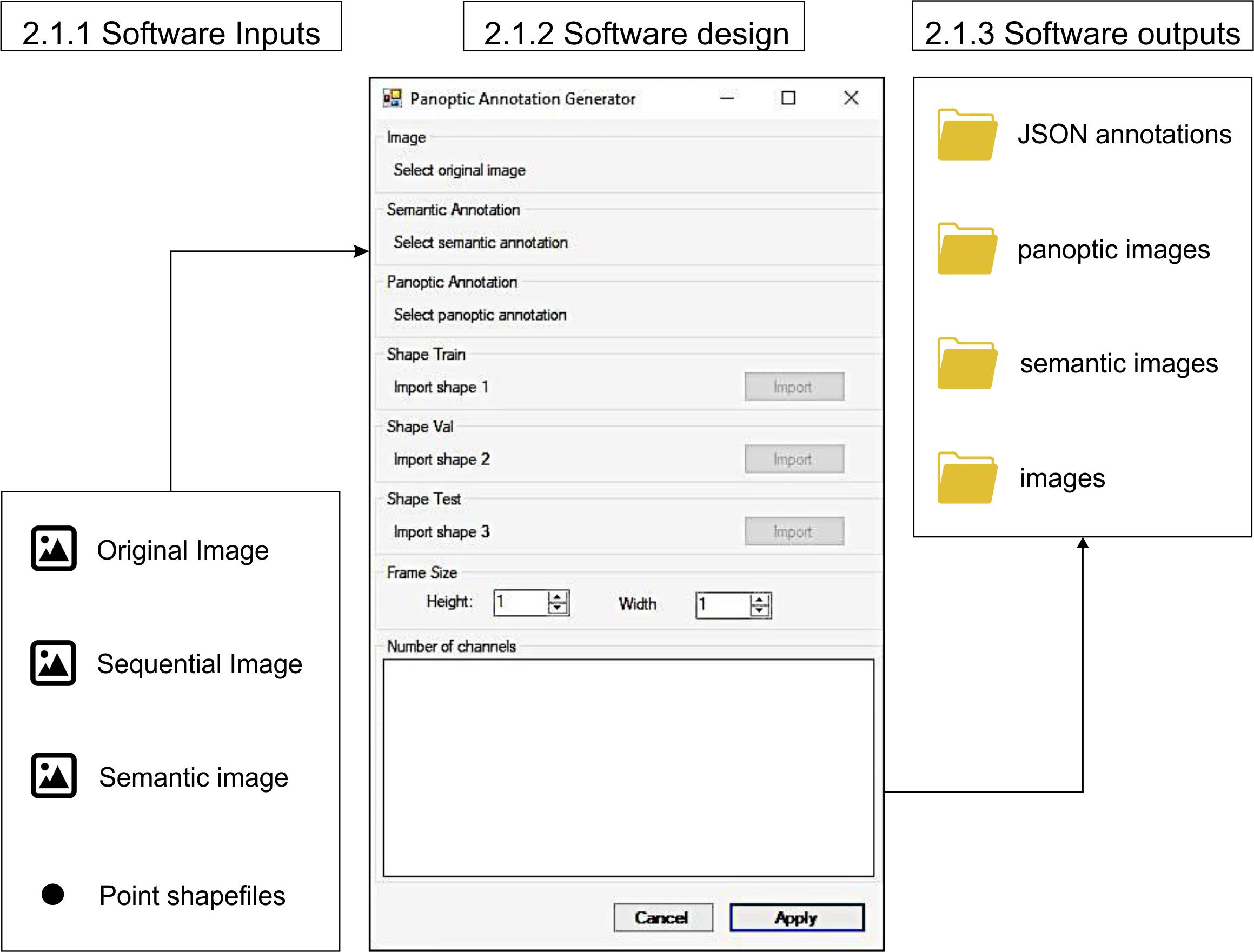}
	\caption{Flowchart of the proposed software to convert data into the panoptic format, including the inputs, design, and outputs.}
	\label{fig:fig6}
\end{figure}

The present study develops a module within the free Abilius software that converts GIS vector data into COCO-compatible annotations widely used in deep learning algorithms (Figure \ref{fig:fig6}) (\url{https://github.com/abilius-app/Panoptic-Generator}). The proposed framework generates samples from vector data in shape format to JSON files in the COCO annotation format, considering the three main segmentation tasks (semantic, instance, and panoptic). The use of GIS databases provides a practical way to expand the free community-maintained datasets, minimizing the time-consuming and challenging process of manually generating large numbers of annotations for different classes of objects. The tool generates annotations for the three segmentation tasks in an end-to-end approach, in which the annotations are ready to use, requiring no intermediary process and reducing labor-intensive work. However, the Detectron2 algorithm requires some changes to receive images with more channels than traditional RGB and use TIFF format instead of conventional JPEG files. This tool was crucial to build the current dataset, but it also applies to other scenarios considering the different DL tasks.

\subsubsection{Software inputs}

\begin{figure}[H]
	\centering %
	\scriptsize %
	\includegraphics[width=0.85\columnwidth]{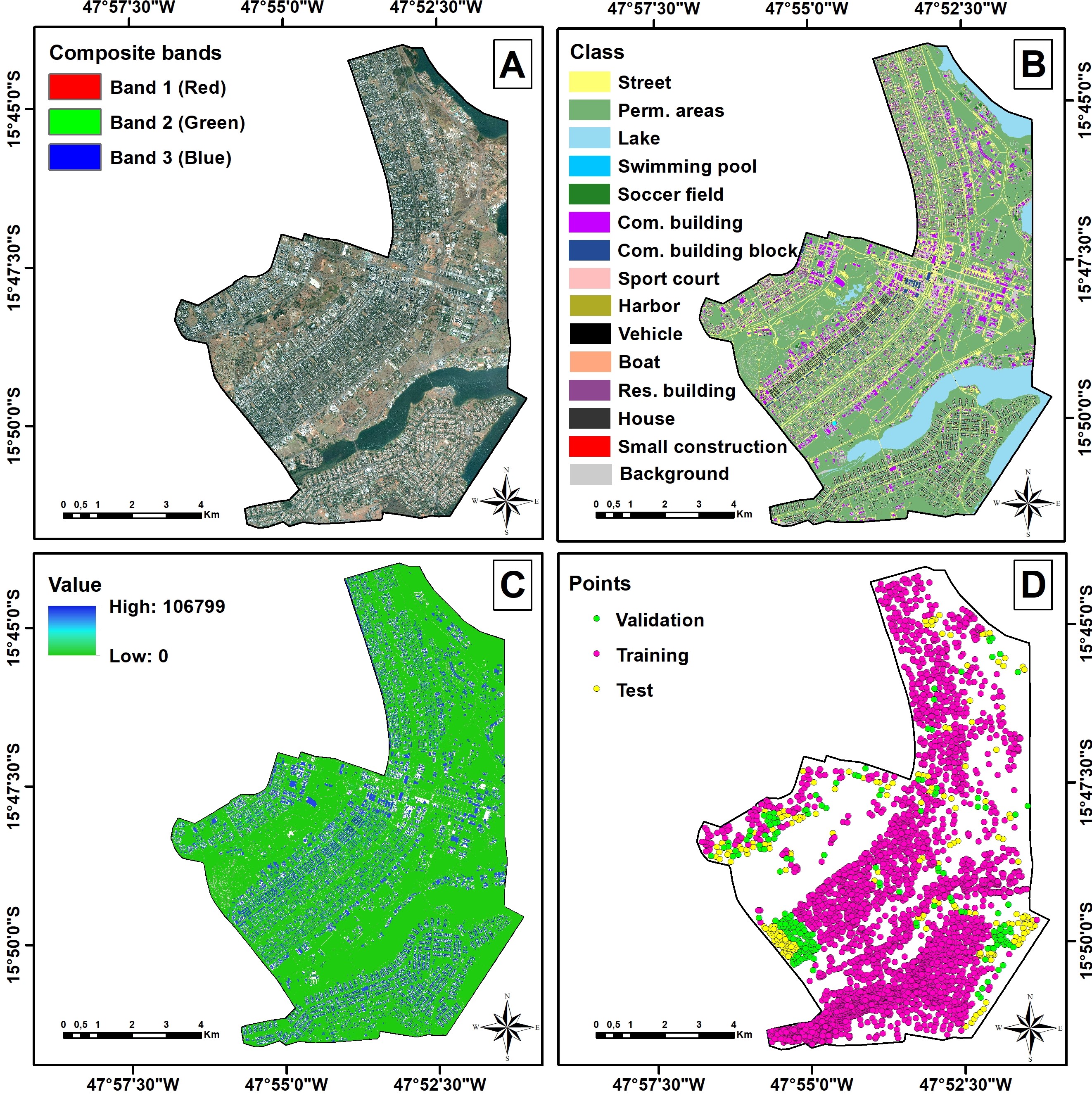}
	\caption{inputs for the software in which (A) is the original image, (B) Semantic image, (C) sequential image, and (D) the point shapefiles for training, validation, and testing.}
	\label{fig:fig7}
\end{figure}

To automatically obtain the semantic, instance, and panoptic annotations, we proposed a novel pipeline with four inputs (considering the georeferenced images in the same system): (a) the original image (Figure \ref{fig:fig7}A); (b) semantic image (Figure \ref{fig:fig7}B); (c) sequential ground truth image (Figure \ref{fig:fig7}C) (each “thing” object has a different value), and (d) the point shapefiles (Figure \ref{fig:fig7}D). The class-agnostic image is a traditional semantic segmentation ground truth, in which each class receives a unique label, easily achieved by converting from polygon to raster in GIS software. The sequential ground truth (which will become the panoptic images) requires a different value for each polygon that belongs to the “thing” categories. First, we grouped all the “stuff” classes since these classes do not need a unique identification. The subsequent “thing” classes receive a unique value to each polygon using sequential values in the attribute table. Moreover, the point shapefiles play a crucial role in generating the DL samples since it uses the point location as the centroid of the frame. Our proposed method using point shapefiles provides the following benefits: (a) more control over the selected data in each set; (b) allows augmenting the training data by choosing points close to each other; and (c) in large images, there are areas with much less relevance, and the user may choose more significant regions to generate the dataset. Apart from the inputs, the user may choose other parameters such as spectral bands and spatial dimensions. Our study used the RGB channels (other applications might require more channels or less depending on the sensor) and 512x512-pixel dimensions.

\subsubsection{Software Design}
Given the raw inputs, the software must crop tiles in the given point shapefile areas. For each point shapefile, it crops all input images considering the point as the centroid, meaning that if the user chooses a tile size of 512x512, the frame will present distance from the centroid of 256-pixels in the up, down, right, and left directions (resulting in a squared frame with 512x512 dimensions). Now, for each 512x512 tile, we must gather the image annotations semantic, instance, and panoptic segmentation tasks, given as follows:

\begin{description}
  \item[$\bullet$] \textbf{Semantic segmentation annotation}: Pixel-wise classification of the entire image with the same spatial dimensions from the original image tiles. Usually, the background (i.e., unlabeled data) has a value of zero. Each class presents a unique value.
  \item[$\bullet$] \textbf{Instance segmentation annotation}: Each object requires a pixel-wise classification, bounding box, and class of each bounding box for each object. Since there is more information when compared to the semantic segmentation approach, most software adopts the COCO annotation format, e.g., Detectron2 \citep{Wu2019Detectron2}. For instance segmentation, the COCO annotation format uses a JSON file requiring for each object the: (a) identification, (b) image identification, (c) category identification (i.e., the label of the class), (d) segmentation (polygon coordinates), (e) area (total number of pixels), (f) bounding box (four coordinates) (\url{https://cocodataset.org/#format-data}).
  \item[$\bullet$] \textbf{Panoptic segmentation annotation}: The panoptic segmentation combines semantic and instance segmentation. It requires a folder with the semantic segmentation images in which all thing classes have zero value. Besides, it requires the instance segmentation JSON file and an additional panoptic segmentation JSON file. The panoptic JSON is very similar to the instance JSON, but considering an identifier named “isthing”, in which the “thing” category is one and “stuff” is zero.
\end{description}

The semantic segmentation data is the most straightforward, and its output cropped tiles are already in the format to apply a semantic segmentation model. Nevertheless, the semantic image plays a crucial role in the instance and panoptic JSON construction. The parameters designed to build the COCO annotation JSONS for instance and panoptic segmentation were the following:

\begin{description}
  \item[$\bullet$] \textbf{Image identification}: Each cropped tile receives an ascending numeration. For example, there are 3,000-point shapefiles in the training set, and the image identifications range from 1 to 3,000.
  \item[$\bullet$] \textbf{Segmentation}: We used the OpenCV C++ library for obtaining all contours in the sequential image. The contour representation is in tuples (x and y). For each distinct value, the proposed software gathers all coordinates separately according to the COCO annotation specifications. The polygon information will only be stored in the instance segmentation JSON, but these coordinates will guide the subsequent bounding box process.
  \item[$\bullet$] \textbf{Bounding box}: Using the polygons obtained in the segmentation process enables the extraction of minimum and maximum points (in the horizontal and vertical directions). There are many possible ways to obtain the bounding box information using four coordinates. However, we used the top-left coordinates associated with the width and height.
  \item[$\bullet$] \textbf{Area}: We apply a loop to count the number of pixels of each different value on the sequential image.
  \item[$\bullet$] \textbf{Category identification}: This is where the segmentation image is so important. The sequential image does not contain any class information (only that each thing class has a different value). For each generated polygon, we extract the category value from the semantic image to use it as the category identification label.
  \item[$\bullet$] \textbf{Object identification}: This method is different for the instance and panoptic JSONS. In the instance JSON, the identification is a sequential ascending value (the last object in the last image will present the highest value, and the first object in the first image will present the lowest value), and it only considers the “thing” classes. In the panoptic JSON, the identification is the same as the object number in sequential order, and it considers “thing” and “stuff” classes.
\end{description}

Apart from these critical parameters, we did not consider the possibility of crowded objects (our data has all separate instances), so the $“is\_crowd”$ parameter is always zero. Moreover, the user must specify which classes are “stuff” or “things”. The sequential input data is an image with single-channel TIFF format transformed in our software to a three-channel PNG image compatible with Detectron2 software, converting from decimal number to base-256.

\subsubsection{Software Outputs}
The software outputs the images and annotations in a COCO dataset structure. The algorithm produces ten folders, an individual folder for annotations in JSON format and three folders for each set of samples (training, validation, and testing) referring to the original image, panoramic annotations, and semantic annotations.
In the training-validation-test split, the training set usually presents most of the data for the purpose of learning the specific task. However, the training set alone is not sufficient to build an effective model since, in many situations, the model overfits the data after a certain point. Thus, the validation set allows tracking the trained model performance on new data while still tuning hyperparameters. The test set is an independent set to evaluate the performance. Table \ref{tab:tab3} lists the number of tiles in each set and the total number of instances. Our proposed conversion software allows overlapping image tiles, which may be valuable in the training data functioning as a data augmentation method. However, this would lead to biased results if applied in the validation and testing sets. In this regard, we used the graphic Buffer analysis tool from the ArcMap software, considering the dimensions generating 512x512 squared buffers to verify that none of the sets were overlapping.

\begin{table}[!h]
\centering
\setlength\extrarowheight{-3pt}
\caption{Data split on the three sets with their respective number of images and instances, in which all images present 512x512x3 dimensions.}
\begin{tabular}{lll}
\hline
 Set          & Number of tiles & Number of instances    \\
 \hline 
 Training     & 3,000           & 102,971                \\
 Validation   & 200             & 9,070                  \\
 Testing      & 200             & 7,237                  \\
 \hline
\end{tabular} 
\label{tab:tab3}
\end{table}

\subsection{Panoptic segmentation model}
\begin{figure}[!h]
	\centering %
	\scriptsize %
	\includegraphics[width=0.9\columnwidth]{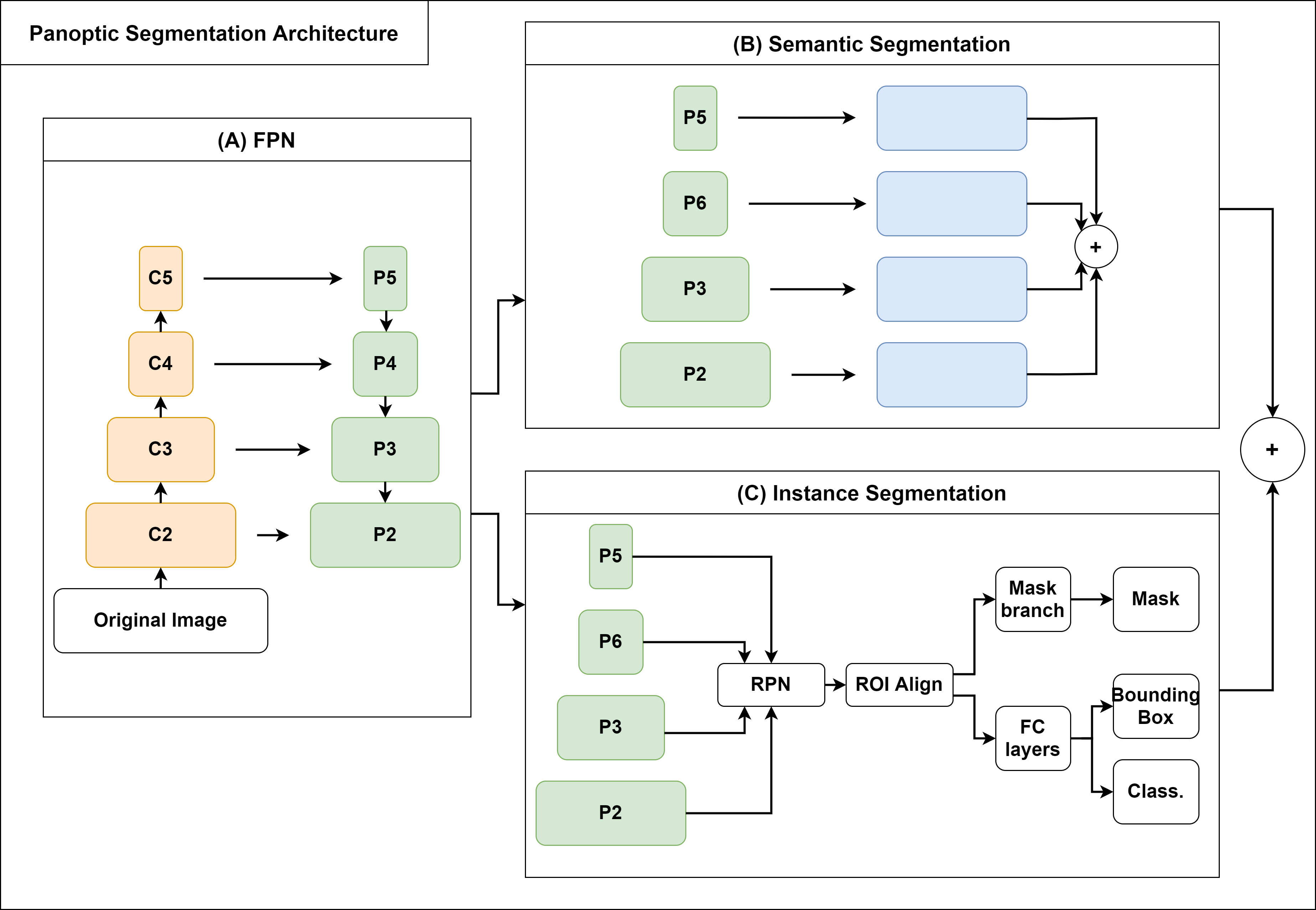}
	\caption{Architecture of the Panoptic Feature Pyramid Network (FPN), with its semantic segmentation (B) and instance segmentation (C) branches.}
	\label{fig:fig8}
\end{figure}

With the annotations in the correct format, the next step was to use panoptic segmentation DL models. Panoptic segmentation networks aim to combine the semantic and instance results using a simple heuristic method \citep{Kirillov2019Panoptic} (Figure \ref{fig:fig8}A). The model presents two branches: semantic segmentation (Figure \ref{fig:fig8}B) and instance segmentation (Figure \ref{fig:fig8}C). Figure \ref{fig:fig8} shows the Panoptic-FPN architecture, which combines the Mask-RCNN for instance predictions and dilates FPN for semantic segmentation. We considered two backbones (ResNet-50 and ResNet-101).

\subsubsection{Semantic segmentation branch}
Semantic segmentation models are the most used among the remote sensing community, mainly because of the good results and simplicity of models and annotation formats. There are a wide variety of architectures such as the U-net \citep{Ronneberger2015U-Net:}, Fully Convolutional Networks (FCN) \citep{Zhang2018Fully}, DeepLab \citep{Chen2018Encoder-Decoder}. The semantic segmentation using the FPN presents some differences when compared to traditional encoder-decoder structures. FPN predictions with different resolutions (P2, P3, P4, P5) are resized to the input image spatial resolution by applying bilinear upsampling, in which the sampling rate is different for each prediction to obtain the same dimensions. The elements present in the “things” category all receive the same label (avoiding problems with the predictions from the instance segmentation branch).

\subsubsection{Instance segmentation branch}
Instance segmentation had a significant breakthrough with the Mask-RCNN \citep{He2020Mask}. This method relies on the extension of Faster-RCNN \citep{Girshick2015Fast}, a detector with two stages: (a) Region Proposal Network (RPN); and (b) box regression and classification for each Region of Interest (ROI) from the RPN. However, aiming to perform pixel-wise segmentation, the Mask-RCNN added a segmentation branch on top of the Faster-RCNN architecture. First, the method applies the RPN on top of different scale predictions (e.g., P2, P3, P4, P5) and proposes several anchor boxes in more susceptible regions. Then, the ROI align procedure standardizes each bounding box dimension (avoiding quantization problems). The last step considers a binary segmentation mask for each object alongside the bounding box with its respective classification.

\subsubsection{Model configurations}
The loss function for the Panoptic-FPN model is the combination of the semantic and instance segmentation losses. The instance segmentation encompasses the bounding box regression, classification, and mask losses. The semantic segmentation uses a traditional cross-entropy loss among the “stuff” categories and a class considering all “thing” categories together.

Regarding the model hyperparameters, we used: (a) stochastic gradient descent (SGD) optimizer, (b) learning rate of 0.0005, (c) 150,000 iterations, (d) five anchor boxes (with sizes 32, 64, 128, 256, and 512), (e) three aspect ratios (0.5, 1, 2), (f) one image per batch. Besides, we trained the model using ImageNet pre-trained weights and unfreezing all layers. Moreover, we evaluated the metrics on the validation set with a period of 1,000 iterations and saved the final model with the highest PQ metric. To avoid overfitting and increase performance (mainly on the small objects), we used three augmentation strategies: (a) random vertical flip (probability chance of 50\%), (b) random horizontal flip (probability chance of 50\%), and (c) resize shortest edge with 640, 672, 704, 736, 768, and 800 possible sizes. The data processing used a computer containing an Intel i7 core and NVIDIA 2080 GPU with 11GB RAM.

\subsection{Model evaluation}
In supervised learning tasks, the accuracy analysis compares the predicted results and the ground truth data. Each task has different ground truth data and, therefore, different evaluation metrics. However, the confusion matrix is a common structure for all tasks, yielding four possible results: true positives (TP), true negatives (TN), false positives (FP), and false negatives (FN). Sections 2.4.1, 2.4.2, and 2.4.3 explain the semantic, instance, and panoptic segmentation metrics, respectively.

\subsubsection{Stuff evaluation}
For semantic segmentation tasks, the confusion matrix analysis is per pixel. The most straightforward metric is the pixel accuracy (pAcc):

\begin{equation}
pAcc=\frac{TP+TN}{TP+TN+FP+FN}
\end{equation}

However, in many cases, the classes are imbalanced, bringing imprecise results. The mean pixel accuracy (mAcc) takes into consideration the number of pixels belonging to each class, performing a weighted average.

Apart from PA, the intersection over union (IoU) is the primary metric for many semantic segmentation studies, mainly because it penalizes the algorithm for FP and FN errors:

\begin{equation}
IoU = \frac{|A \cap  B|}{|A \cup B|} = \frac{TP}{TP+FP+FN}
\end{equation}

In which:
$A \cap B$: the area of intersection;
$A \cup B$: the area of union.

For a more general understanding of this metric, we may use the mean IoU (mIoU), which is the average IoU of all categories or the frequency weighted IoU (fwIoU) which is the weighted average of each IoU considering the frequency of each class.

\subsubsection{Thing evaluation}
Instance segmentation metrics take into consideration both the bounding box predictions and the mask quality. The most common approach to instance segmentation problems uses standard COCO metrics \citep{Bolya2020YOLACT++:, Cai2018Cascade, Gao2021Res2Net:, He2020Mask, Huang2019Mask}. The primary metric in evaluation is the average precision (AP) \citep{Lin2014Microsoft}, also known as the area under the precision-recall curve:

\begin{equation}
AP=\ \int_{0}^{1}Precision\left(Recall\right)dRecall,
\end{equation}

in which:
\begin{equation}
Precision=\ \frac{TP}{TP+FP}
\end{equation}
\begin{equation}
Recall=\ \frac{TP}{TP+FP}
\end{equation}

Moreover, the COCO AP metrics consider different IoU thresholds from 0.5 to 0.95 with 0.05 steps, which is useful to measure the quality of the bounding boxes compared to the original image. The secondary metrics consider specific IoU thresholds: AP$_{50}$ and AP$_{75}$, which use IoU values of 0.5 and 0.75, respectively. Besides, the evaluation also includes objects of different sizes in three categories (AP$_{S}$, AP$_{M}$, and AP$_{L}$): (1) small objects (area $<$ 32$^2$ pixels); (2) medium objects (32$^2$ pixels $<$ area $<$ 96$^2$ pixels); and (3) large objects (area $<$ 96$^2$ pixels).

\subsubsection{Panoptic evaluation}
The Panoptic Quality (PQ) is the primary metric for evaluating the PS task \citep{Gao2021Res2Net:, Kirillov2019Panoptic, Mohan2021EfficientPS:}, and it is the current metric for the COCO panoptic task challenge, being defined by: 

\begin{equation}
PQ=\frac{\sum_{\left(p,g\right)\in TP}\ IoU\left(pred,GT\right)}{\left|TP\right|+\frac{1}{2}\left|FP\right|+\frac{1}{2}\left|FN\right|}
\end{equation}

In which:
pred: the DL prediction;
GT: the ground truth.

The expression above enables two metrics: Segmentation Quality (SQ) related to the segmentation task and the Recognition Quality (RQ) to object detection, considering the following equations:

\begin{equation}
SQ=\frac{\sum_{\left(p,g\right)\in TP}\ IoU(pred,GT)}{\left|TP\right|}\ 
\end{equation}

\begin{equation}
RQ=\frac{TP}{\left|TP\right|+\frac{1}{2}\left|FP\right|+\frac{1}{2}|FN|}
\end{equation}

\section{Results}
\subsection{Metrics}
The metrics section presents (3.1.1) semantic segmentation metrics, (3.1.2) instance segmentation metrics, and (3.1.3) panoptic segmentation metrics. The semantic segmentation metrics are related to the “stuff” classes in a per-pixel analysis. The instance segmentation classes relate to the “thing” classes using traditional object detection metrics, such as the AP. The panoptic segmentation metrics englobes both types of features.

\subsubsection{Semantic segmentation results}
Table \ref{tab:tab4} lists the general metrics for the three “stuff” categories (street, permeable area, and lake), considering the mIoU, fwIoU, mAcc, and pAcc for the Panoptic-FPN model with the ResNet-50 and ResNet-101 backbones. The validation and test results were very similar, in which the R101 backbone presented slightly better results among all metrics. In the validation and test sets, the metric with the most considerable difference between the ResNet-50 and ResNet-101 backbones was the IoU (0.514 and 1.484 difference in the validation and test set, respectively).

\begin{table}[!h]
\centering
\setlength\extrarowheight{-3pt}
\caption{Mean Intersection over Union (mIoU), frequency weighted (fwIoU), mean accuracy (mAcc), and pixel accuracy (pAcc) results for semantic segmentation in the BSB Aerial Dataset validation and test sets.}
\begin{tabular}{lllll}
\hline
 Backbone     & mIoU & fwIoU   & mAcc   & pAcc    \\
 \hline
 \multicolumn{5}{c}{Validation set}              \\
 \hline
 R50         & 92.129 & 92.865 & 95.643 & 96.271 \\
 R101        & 92.643 & 93.241 & 95.769 & 96.485 \\
 Difference  & 0.514  & 0.376  & 0.126  & 0.214  \\
 \hline
 \multicolumn{5}{c}{Test set}                    \\
 \hline
  R50        & 92.381 & 93.404 & 95.772 & 96.573 \\
 R101        & 93.865 & 94.472 & 96.339 & 97.148 \\
 Difference  & 1.484  & 1.068  & 0.567  & 0.575  \\
 \hline
\end{tabular} 
\label{tab:tab4}
\end{table}

Table \ref{tab:tab5} lists the accuracy results of each "stuff" class for the validation and test sets. In addition to the three stuff classes (lake, permeable area, and street), the analysis creates another class merging the "thing" classes (we defined it as "all things"). Some samples have a single-class predominance, such as lake and permeable area, increasing the accuracy metric due to the high proportion of correctly classified pixels. The "lake" class presented the highest IoU for the validation (97.1\%) and test (97.8\%) sets, mainly because it presents very distinct characteristics from all other classes in the dataset. The permeable area achieves a slightly lower accuracy (IoU of 95.384 for validation and 96.275 for the test) than the lake class because it encompasses many different intraclass features (e.g., trees, grass, earth, sand). The "street" class, widely studied in remote sensing, presented an IoU of 88\% and 90\% for validation and test. These IoU values are significant considering the difficulty of street mapping even by visual interpretation due to the high interference of overlapping objects (e.g., cars, permeable areas, undefined elements) and the challenges with shaded areas.

\begin{table}[!h]
\centering
\setlength\extrarowheight{-3pt}
\caption{ Segmentation metrics (Intersection over Union (IoU) and Accuracy (Acc)) for each “stuff” classes in the BSB Aerial dataset validation and test sets considering the ResNet101 (R101), ResNet50 (R50) backbones, and their difference (R101-R50).}
\begin{tabular}{l|ll|ll|ll}
\hline
\multirow{2}{*}{Category} & \multicolumn{2}{c}{R101} & \multicolumn{2}{c}{R50} & \multicolumn{2}{c}{Difference} \\      
 \cline{2-7}
            & IoU    & Acc    & IoU    & Acc   & IoU    & Acc    \\
 \hline
 \multicolumn{7}{c}{Validation set}                                      \\
 \hline
 All things         & 89.962 & 95.060 & 89.402 & 94.882 & 0.56  & 0.178   \\
 Street             & 88.079 & 91.773 & 86.933 & 91.799 & 1.146 & -0.026  \\
 Permeable Area     & 95.384 & 98.090 & 95.286 & 97.786 & 0.098 & 0.304   \\
 Lake               & 97.148 & 98.153 & 96.993 & 98.105 & 0.155 & 0.048   \\
 \hline
 \multicolumn{7}{c}{Test set}                                             \\
 \hline
 All things         & 90.718 & 94.563 & 89.142 & 93.041 & 1.576 & 1.522   \\
 Street             & 90.607 & 93.600 & 89.129 & 93.844 & 1.478 & -0.244  \\
 Permeable Area     & 96.275 & 98.775 & 95.559 & 98.120 & 0.716 & 0.655   \\
 Lake               & 97.859 & 98.459 & 95.665 & 98.013 & 2.194 & 0.446   \\
 \hline
\end{tabular}
\label{tab:tab5}
\end{table}

The R101 backbone presented better IoU results for all categories. The most significant difference was the street category in the validation set (1.146) and the lake in the test set (2.194). The R50 backbone presented a higher value for the street class in the validation (0.026) and test sets (0.244). Since the balancing of the classes is not even, the IoU provides more insightful results when compared to the accuracy.

\subsubsection{Instance segmentation results}
Table \ref{tab:tab6} lists the results for the standard COCO metrics (AP, AP$_{50}$, AP$_{75}$, AP$_{S}$, AP$_{M}$, and AP$_{L}$) for the “thing” classes, considering the bounding box (Box) and segmentation mask (mask), from the two backbones (ResNet-101 (R101) and ResNet-50 (R50)). The validation and test results were very similar to those occurring in the “stuff” classes. However, the primary metric (AP) differences among the two backbones (R101 – R50) were more considerable in the test set regarding the box metrics, with a difference of nearly 1.6\%. The R101 backbone had higher values in almost all derived metrics, except for the AP75 box metric in the validation set and the APmedium in the test set.

\begin{table}[!h]
\centering
\setlength\extrarowheight{-3pt}
\caption{COCO metrics for the “thing” categories in the BSB Aerial Dataset validation set considering two backbones (ResNet-101 (R101) and ResNet-50 (R50)) and their difference (R101 – R50).}
\begin{tabular}{l|l|l|l|l|l|l|l}
\hline
 Backbone        & Type & AP     & $AP_{50}$   & $AP_{75}$   & $AP_{S}$  & $AP_{M}$ & $AP_{L}$  \\
 \hline
 \multicolumn{8}{c}{Validation set}                                                                          \\
 \hline
 \multirow{2}{*}{R101} & Box   & 47.266 & 69.351 & 50.206 & 26.154   & 51.667    & 55.680                            \\
                       & Mask  & 45.379 & 68.331 & 50.917 & 24.064   & 49.490   & 57.882                            \\
                \hline
 \multirow{2}{*}{R50}  & Box   & 45.855 & 68.258 & 51.351 & 25.806   & 49.732   & 48.678                            \\
                       & Mask  & 42.850 & 68.553 & 48.863 & 21.213   & 47.686    & 47.040                            \\
             \hline
 \multirow{2}{*}{Difference} & Box   & 1.411  & 1.093  & -1.145 & 0.348    & 1.935    & 6.993                             \\
                             & Mask  & 2.529  & 2.778  & 2.054  & 2.851    & 1.804    & 10.842                            \\
 \hline
 \multicolumn{8}{c}{Test set}                                                                                \\
 \hline
  \multirow{2}{*}{R101} & Box   & 47.691 & 67.096 & 52.552 & 28.920   & 49.795   & 57.446                            \\
                        & Mask  & 44.211 & 65.271 & 49.394 & 25.016   & 49.377   & 58.311                            \\
            \hline
 \multirow{2}{*}{R50}   & Box   & 44.642 & 64.306 & 50.727 & 28.636   & 49.881   & 53.298                            \\
                        & Mask  & 41.933 & 62.821 & 47.640 & 23.631   & 50.027   & 52.204                            \\
            \hline
 \multirow{2}{*}{Difference} & Box   & 3.049  & 2.790  & 1.825  & 0.284    & -0.086   & 4.148                             \\
                             & Mask  & 2.278  & 2.450  & 1.754  & 1.385    & -0.650   & 6.107                            \\
 \hline
\end{tabular}
\label{tab:tab6}
\end{table}

Although the overall metrics showed better performance for the R101 backbone, the analysis by class presents some classes with slightly better results for the R50 backbone (Table \ref{tab:tab7}). In the validation set, five of the eleven classes had higher values in the ResNet-50 backbone (harbor, boat, soccer field, house, and small construction). This effect was less frequent in the test set, showing only the boat class with superiority of the ResNet-50 backbone in the box metric and three classes (swimming pool, boat, and commercial building) in the mask metric.

\begin{table}[!h]
\centering
\setlength\extrarowheight{-5pt}
\caption{AP metrics for bounding box and mask per category considering the “thing” classes in the BSB Aerial Dataset validation set for the ResNet101 (R101) and ResNet50 (R50) backbones and their difference (R101-R50).}
\begin{tabular}{l|ll|ll|ll}
\hline
\multirow{2}{*}{Category} & \multicolumn{2}{c}{R101} & \multicolumn{2}{c}{R50} & \multicolumn{2}{c}{Difference} \\      
 \cline{2-7}
 & Box AP & Mask AP & Box AP & Mask AP & Box AP & Mask AP    \\
 \hline
 \multicolumn{7}{c}{Validation set}                                           \\
 \hline
Swimming   pool        & 55.495 & 53.857 & 53.121 & 51.974 & 2.374  & 1.883  \\
Harbor                 & 37.137 & 21.079 & 39.415 & 24.300 & -2.278 & -3.221 \\
Vehicle                & 55.616 & 56.573 & 54.568 & 55.893 & 1.048  & 0.680   \\
Boat                   & 30.582 & 36.216 & 35.329 & 37.265 & -4.747 & -1.049 \\
Sports   court         & 56.681 & 55.193 & 46.906 & 42.494 & 9.775  & 12.699 \\
Soccer   field         & 34.866 & 39.569 & 39.619 & 41.767 & -4.753 & -2.198 \\
Com.   building        & 32.114 & 31.799 & 28.592 & 28.471 & 3.522  & 3.328  \\
Com.   building block  & 66.283 & 63.192 & 52.149 & 47.606 & 14.134 & 15.586 \\
Residential   building & 67.046 & 57.615 & 63.512 & 54.312 & 3.534  & 3.303  \\
House                  & 57.555 & 56.697 & 59.907 & 57.470 & -2.352 & -0.773 \\
Small   construction   & 26.550 & 27.381 & 31.284 & 29.800 & -4.734 & -2.419 \\
 \hline
 \multicolumn{7}{c}{Test set}                                                  \\
 \hline
Swimming   pool        & 53.561 & 50.044 & 51.546 & 50.520 & 2.015  & -0.476 \\
Harbor                 & 42.429 & 22.837 & 31.409 & 17.270 & 11.02  & 5.567  \\
Vehicle                & 56.371 & 57.689 & 55.695 & 57.311 & 0.676  & 0.378  \\
Boat                   & 26.190 & 31.210 & 30.698 & 34.875 & -4.508 & -3.665 \\
Sports   court         & 46.018 & 45.515 & 40.566 & 40.672 & 5.452  & 4.843  \\
Soccer   field         & 46.279 & 45.831 & 36.832 & 33.886 & 9.447  & 11.945 \\
Com.   building        & 42.516 & 37.709 & 41.145 & 40.265 & 1.371  & -2.556 \\
Com.   building block  & 70.971 & 67.465 & 69.341 & 63.679 & 1.63   & 3.786  \\
Residential   building & 54.829 & 47.397 & 51.774 & 44.640 & 3.055  & 2.757  \\
House                  & 62.395 & 59.886 & 57.861 & 58.396 & 4.534  & 1.490   \\
Small   construction   & 26.046 & 20.740 & 24.202 & 19.746 & 1.844  & 0.994 \\
 \hline
\end{tabular}
\label{tab:tab7}
\end{table}

\subsubsection{Panoptic segmentation results}
Table \ref{tab:tab8} lists the results for the panoptic segmentation metrics (PQ, SQ, and RQ), which are the main metrics for evaluating this task. In hand with the previous “stuff” and “thing” results, the ResNet-101 backbone presented the best metrics in most cases, except for the RQ$_{stuff}$ in the validation set and the SQ$_{things}$ in the test set. Overall, the main metric for analysis (PQ) had nearly a 2\% difference among the backbones. The low discrepancies among the different architectures suggest that in situations with lower computational power, the usage of a lighter backbone still presents close enough results.

\begin{table}[!h]
\centering
\setlength\extrarowheight{-5pt}
\caption{COCO metrics for panoptic segmentation in the BSB Aerial Dataset validation and test sets considering the Panoptic Quality (PQ), Segmentation Quality (SQ), and Recognition Quality (RQ).}
\begin{tabular}{l|l|l|l|l}
\hline
Backbone & Type & PQ & SQ & RQ \\
\hline
\multicolumn{5}{c}{Validation Set}   \\
\hline
\multirow{3}{*}{R-101}      & All    & 65.296 & 85.104 & 76.229 \\
                            & Things & 59.783 & 82.876 & 71.948 \\
                            & Stuff  & 85.508 & 93.272 & 91.925 \\
                            \hline
\multirow{3}{*}{R-50}       & All    & 63.829 & 84.886 & 74.550 \\
                            & Things & 57.958 & 82.777 & 69.674 \\
                            & Stuff  & 85.354 & 92.617 & 92.432 \\
                            \hline
\multirow{3}{*}{Difference} & All    & 1.467  & 0.218  & 1.679  \\
                            & Things & 1.825  & 0.099  & 2.274  \\
                            & Stuff  & 0.154  & 0.655  & -0.507 \\
                            \hline
\multicolumn{5}{c}{Test Set}                                    \\
\hline
\multirow{3}{*}{R-101}      & All    & 64.979 & 85.378 & 75.474 \\
                            & Things & 58.354 & 83.171 & 69.997 \\
                            & Stuff  & 89.272 & 93.468 & 95.558 \\
                            \hline
\multirow{3}{*}{R-50}       & All    & 62.230 & 85.315 & 72.179 \\
                            & Things & 55.239 & 83.344 & 65.956 \\
                            & Stuff  & 87.864 & 92.540 & 94.998 \\
                            \hline
\multirow{3}{*}{Difference} & All    & 2.749  & 0.063  & 3.295  \\
                            & Things & 3.115  & -0.173 & 4.041  \\
                            & Stuff  & 1.408  & 0.928  & 0.560  \\
                            \hline
\end{tabular}
\label{tab:tab8}
\end{table}

\subsection{Visual Results}
\begin{figure}[H]
	\centering %
	\scriptsize %
	\includegraphics[width=1\columnwidth]{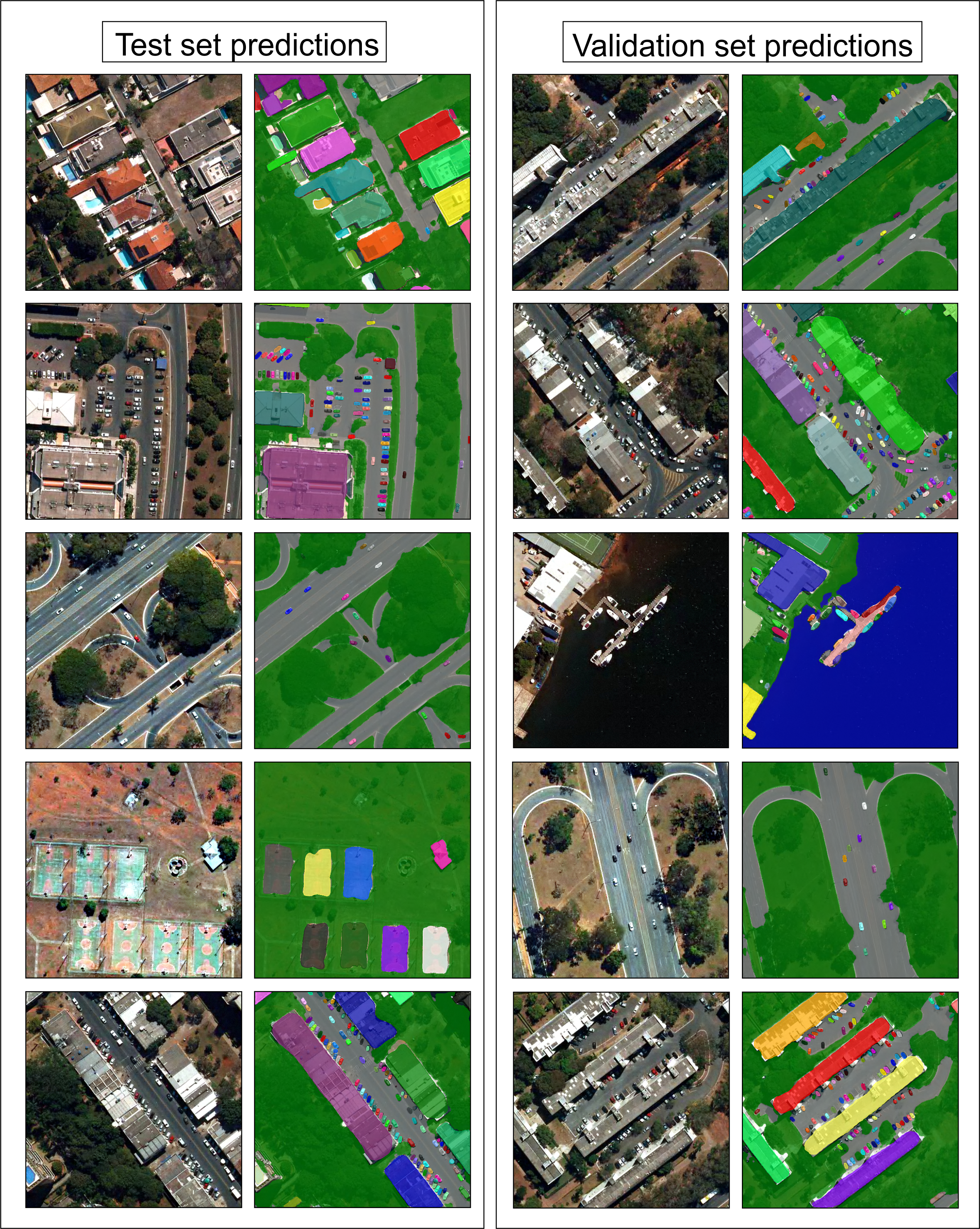}
	\caption{Five examples of predictions on the test and validation sets including the original 512x512 Red, Green, and Blue (RGB) image and the corresponding panoptic prediction.}
	\label{fig:fig9}
\end{figure}

Figure \ref{fig:fig9} shows five test and validation samples, including the original images and predictions from the Panoptic-FPN model using the ResNet-101 backbone. The results demonstrate a coherent urban landscape segmentation, visually integrating countable objects (things) and amorphous regions (things) in an enriching perspective toward real-world representation. As shown in the metrics section, the results show no evident discrepancies in the validation and test data, demonstrating very similar visual results in both sets. The segmented images show the high ability to visually separate the different instances, even in crowded situations like cars in parking lots. Furthermore, the “stuff” classes are very well delineated, showing little confusion among the street, permeable areas, and lake classes. The set of established classes allows a good representation of the urban landscape elements, even considering some class simplifications. Therefore, panoptic segmentation congregates multiple competencies in computer vision for the satellite imagery interpretation in a single structure.

\section{Discussion}
The panoptic segmentation task imposes new challenges in the formulation of algorithms and database structures, covering particularities of both object detection and semantic segmentation. Therefore, panoptic segmentation establishes a unified image segmentation approach, which changes digital image processing and requires new annotation tools and extensive and adapted datasets. In this context, this research innovates by developing a panoptic data annotation tool, establishing a panoptic remote sensing dataset, and being one of the first evaluations of the use of panoptic segmentation in urban aerial images.

\subsection{Annotation tools for remote sensing}
Many software annotation tools are available online, e.g., LabelMe \citep{Russell2008LabelMe:}. Nevertheless, those tools have problems with satellite image data because of large sizes and other singularities that are uncommon in the traditional computer vision tasks: (a) image format (i.e., satellite imagery is often in GeoTIFF, whereas traditional computer vision uses PNG or JPEG images), (b) georeferencing, and (c) compatibility with polygon GIS data.  The remote sensing field made use of GIS software long before the rise of DL. With that said, there are extensive collections of GIS data (urban, agriculture, change detection) that other researchers could apply DL models. However, vector-based GIS data requires modifications to use DL models. Thus, we proposed a conversion tool from GIS data that automatically crops image tiles with their corresponding polygon vector data stored in shapefile format to panoptic, instance, and semantic annotations. The proposed tool is open access and works independently, without the need to use proprietary programs such as LabelRS developed by ArcPy and dependent on ArcGis \citep{Li2021LabelRS:}. Besides, our proposed pipeline and software enable the users to choose many samples for training, validation, and testing in strategic areas using point shapefiles. This method of choosing samples presents a huge benefit compared to methods such as sliding windows for image generation. Finally, our software enables the generation of the three segmentation tasks (instance, semantic, and panoptic), allowing other researchers to exploit the field of desire.

\subsection{Datasets}
Most transfer learning applications use trained models from extensive databases such as the COCO dataset. Nevertheless, remote sensing images present characteristics that may not yield the most optimal results using traditional images. These images contain diverse targets and landscapes, with different geometric shapes, patterns, and textural attributes, representing a challenge for automatic interpretation. Therefore, the effectiveness of training and testing depends on accurately annotated ground truth datasets, which requires much effort into building large remote sensing databases with a significant variety of classes. Furthermore, the availability of open access encourages new methods and applications, as seen in other computer vision tasks.

Long et al. \citep{Lin2019Improving} performed a complete review of remote sensing image datasets for deep learning methods, including tasks of scene classification, object detection, semantic segmentation, and change detection. In this recent review, there is no database for panoptic segmentation, which demonstrates a knowledge gap. Most datasets consider limited semantic categories or target a specific element, such as building \citep{Benedek2012Building, Ji2019Fully, Etten2018SpaceNet:}, vehicle \citep{Drouyer2020VehSat:, Lin2020VAID:, Zeng2021UAVData:}, ship \citep{Hou2020FUSAR-Ship:, Huang2018OpenSARShip:, Wei2020HRSID:}, road \citep{Das2011Use, Maggiori2017Can}, among others. Regarding available remote sensing datasets for various urban categories, one of the main is the iSAID \citep{Zamir2019iSAID:}, with 2,806 aerial images distributed in 15 different classes, for instance segmentation and object detection tasks.

The scarcity of remote sensing databases with all cityscape elements makes mapping difficult due to highly complex classes, numerous instances, and mainly intraclass and interclass elements commonly neglected. Adopting the panoptic approach allows us to relate the content of interest and the surrounding environment, which is still little explored. Therefore, organizing large datasets into panoptic categories is a key alternative to mapping complex environments such as urban systems that are not reached even with enriched semantic categories. 

The proposed BSB Aerial Dataset contains 3,400 images (3,000 for training, 200 for validation, and 200 for testing) with 512x512 dimensions containing fourteen common urban classes. This dataset simplified some urban classes, such as sports courts instead of tennis courts, soccer fields, and basketball courts. Moreover, our dataset considers three “stuff” classes, widely represented in the urban setting, such as roads. The availability of data and the need for periodic mapping of urban infrastructure by the government allows for the constant improvement of this database. Besides, the dataset aims to trigger other researchers to exploit this task thoroughly. 

\subsection{Difficulties in the Urban Setting}
\begin{figure}[H]
	\centering %
	\scriptsize %
	\includegraphics[width=1\columnwidth]{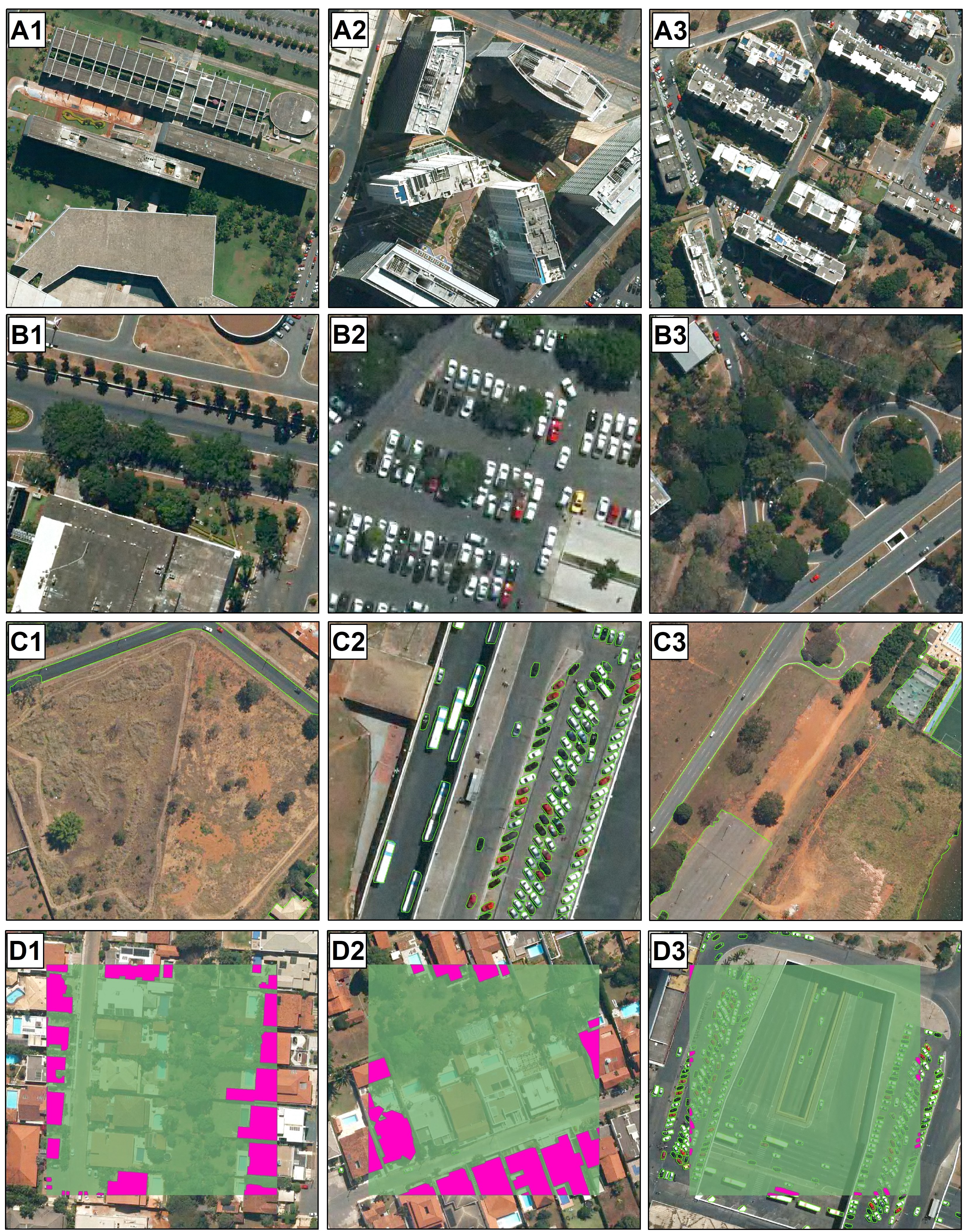}
	\caption{Three examples of: (1) shadow areas (A1, A2, and A3), (2) occluded objects (B1, B2, B3), (3) class categorization (C1, C2, and C3), and (4) edge problem on the image tiles (D1, D2, and D3).}
	\label{fig:fig10}
\end{figure}

Although this study shows a promising field in remote sensing with a good capability of identifying “thing” and “stuff” categories simultaneously, we observed four main difficulties in image annotation and possible results in the urban setting (Figure \ref{fig:fig10}): (1) shadows, (2) occlusion objects, (3) class categorization, and (4) edge problem on the image tiles. Shadows entirely or partially obstruct the light and occur under diverse conditions from the different objects (e.g., cloud, building, mountain, and trees), requiring well-established ground rules to obtain consistent annotations. Therefore, the shadow presence is a source of confusion and misclassification, reducing image quality for visual interpretation and segmentation and, consequently, negatively impacting the accuracy metrics \citep{Wang2017Automatic} (Figure \ref{fig:fig10}A1, \ref{fig:fig10}A2, and \ref{fig:fig10}A3). Specifically, urban landscapes have a high proportion of areas covered by shadows due to the high density of tall objects. Therefore, urban zones aggravate the interference of shadows, causing semantic ambiguity and incorrect labeling, which is a challenge in remote sensing studies \citep{Lin2019Improving, Liu2018ERN:}. Deep learning methods tend to minimize shading effects, but errors occur in very low light locations. Another fundamental problem in computer vision is the occlusion that impedes object recognition in satellite images. Commonly, there are many object occlusions in the urban landscape, such as vehicles partially covered by trees and buildings, making their identification difficult even for humans (Figure \ref{fig:fig10}B1, \ref{fig:fig10}B2, and \ref{fig:fig10}B3). 

Like the occlusion problem, the objects that rely on the tile edges may present an insufficient representation. In monothematic studies, the authors may design the dataset to avoid this problem. However, for the panoptic segmentation task, which aims for an entire scene pixel-wise classification, some objects will be partial representation no matter how large we choose the image tile (Figure \ref{fig:fig10}D1, \ref{fig:fig10}D2, and \ref{fig:fig10}D3). Our proposed annotation tool enables the authors to select each tile's exact point, which gives data generation autonomy to avoid very few representations (even though the problem will still be present). By choosing large image tiles, the percentual representation of edge objects will be lower and tends to have a smaller impact on the model and accuracy metrics but increasing the image tile also requires more computational power.

Finally, the improvement of urban classes in the database is ongoing work. This research sought to establish general and representative classes, but the advent of new categories will allow for more detailed analysis according to research interests. For example, our vehicle class encompasses buses, small cars, and trucks, and our permeable area class contains bare ground, grass, and trees as shown in Figures \ref{fig:fig10}C1, \ref{fig:fig10}C2, and \ref{fig:fig10}C3.

\subsection{Panoptic segmentation task}
This study brings an analysis comparing three approaches which all contain some advantages and disadvantages. The building individualization, especially when crowded, may be challenging even for annotation specialists, consisting of one of the main obstacles in the instance and panoptic segmentation. On the other hand, semantic segmentation does not aim to predict individual categories for different instances, bringing better accuracy at the cost of not individualizing targets. Therefore, the segmentation approach must consider the research objective. If the goal is to estimate the area of buildings, semantic segmentation might be the best option. In contrast, instance segmentation might be more appropriate if the objective is to estimate buildings individually and estimate the average building size.

Moreover, panoptic segmentation brings advantages for a more thorough scene understanding, enabling the usage of both thing and stuff categories. In this regard, the existing panoptic segmentation studies using novel datasets in remote sensing do not benefit from this method. \cite{Khoshboresh-Masouleh2021Building} applies a panoptic segmentation model for building change detection, which only contains a single class, and \cite{garnot2021panoptic} use different instances for crop detection but without the usage of different backgrounds. The approach is very similar to an instance segmentation method using the Mask-RCNN model since the only considered “stuff” class is the background class. In our method, apart from eleven “thing” classes, we use three well-delineated “stuff” classes (apart from the background) very present in the urban setting.

\section{Conclusions}
The application of panoptic, instance and semantic segmentation often depends on the desired outcome of a research or industry application. Nevertheless, a research gap in the remote sensing community is the lack of studies addressing panoptic segmentation, one of the most powerful techniques. The present research proposed an effective solution for using this unexplored and powerful method in remote sensing by: (a) providing a large dataset (BSB aerial dataset) containing 3,400 images with 512x512 pixel dimensions in the COCO annotation format and fourteen classes (eleven "thing" and three "stuff" categories), being suitable for testing new DL models, (b) providing a novel pipeline and software for easily generating panoptic segmentation datasets in a format that is compatible with state-of-the-art software (e.g., Detectron2), and (c) leveraging and modifying structures in the DL models for remote sensing applicability, and (d) making a complete anaylisis of different metrics and evaluating difficulties of this task in the urban setting. One of the main challenges for preparing a panoptic segmentation model is the image format, which is still not well documented. Thus, we proposed an automatic converter from GIS data to panoptic, instance, and semantic segmentation formats. GIS data was widespread even before the DL rise, and the number of datasets that could benefit from our method is enormous. Besides, our tool allows the users to choose the exact points in large images to generate the DL samples using point shapefiles, which brings more autonomy to the studies and allows better data choosing. We believe that this work may increase other studies on the panoptic segmentation task with the BSB Aerial Dataset and the annotation tool and the baselines comparisons using well-documented software (Detectron2). Moreover, we evaluated the Panoptic-FPN model using two backbones (ResNet-101 and ResNet-50), showing promising metrics for this method's usage in the urban setting. Therefore, this research shows an effective annotation tool, a large dataset for multiple tasks, and their application on some non-trivial models.

\section*{Data Availability Statement}
The data in this research will be made publicly available upon publication of this paper. Meanwhile, the source code and data may be shared with other researchers upon reasonable request to the corresponding author.

\section*{Declaration of competing interest}
The authors declare that they have no known competing financial interests or personal relationships that could have appeared to influence the work reported in this paper.

\section*{Acknowledgments}
The authors are grateful for financial support from CNPq fellowship (Osmar Abílio de Carvalho Júnior, Renato Fontes Guimarães, and Roberto Arnaldo Trancoso Gomes). Special thanks are given to the research group of the Laboratory of Spatial Information System of the University of Brasilia for technical support.

%Bibliography
\clearpage %Put references in their own page
\bibliographystyle{elsevier-model5-names}\biboptions{authoryear}
\bibliography{bibliography.bib}

\end{document}